\pdfoutput=1

\documentclass[11pt]{article}

\usepackage{acl}

\usepackage{times}
\usepackage{latexsym}

\usepackage[T1]{fontenc}

\usepackage[utf8]{inputenc}

\usepackage{microtype}

\usepackage{inconsolata}
\usepackage{hyperref}
\usepackage{url}
\usepackage{xcolor}
\usepackage{color}
\usepackage{colortbl}
\usepackage{booktabs}
\usepackage{graphicx}
\usepackage{dsfont}
\usepackage{wrapfig}
\usepackage{subfig}
\usepackage{overpic}
\usepackage{color}
\usepackage{verbatim}
\usepackage{amsmath} 
\usepackage{multirow} 
\usepackage{wrapfig}
\usepackage[title]{appendix}
\title{EmbSpatial-Bench: Benchmarking Spatial Understanding for Embodied Tasks with Large Vision-Language Models}

\author{Mengfei Du\textsuperscript{${1}$\thanks{Equal contribution}}, Binhao Wu\textsuperscript{${1}$\footnotemark[1]}, Zejun Li\textsuperscript{$1$}, Xuanjing Huang\textsuperscript{$2$}, Zhongyu Wei\textsuperscript{${1}$\thanks{Corresponding author}}
\\
\textsuperscript{1}School of Data Science, Fudan University, China\\
\textsuperscript{2}School of Computer Science, Fudan University, China\\
\texttt{\{mfdu22, bhwu22\}@m.fudan.edu.cn}\\
\texttt{\{zejunli20, xjhuang, zywei\}@fudan.edu.cn}\\
}

\begin{document}
\maketitle
\begin{abstract}
The recent rapid development of Large Vision-Language Models (LVLMs) has indicated their potential for embodied tasks.
However, the critical skill of spatial understanding in embodied environments has not been thoroughly evaluated, leaving the gap between current LVLMs and qualified embodied intelligence unknown. 
Therefore, we construct EmbSpatial-Bench, a benchmark for evaluating embodied spatial understanding of LVLMs. 
The benchmark is automatically derived from embodied scenes and covers 6 spatial relationships from an egocentric perspective. 
Experiments expose the insufficient capacity of current LVLMs (even GPT-4V). 
We further present EmbSpatial-SFT, an instruction-tuning dataset designed to improve LVLMs' embodied spatial understanding. 

\end{abstract}

\section{Introduction}

Embodied AI is the frontier direction of general-purpose AI systems, requiring intelligent agents to understand instructions, perceive physical environments, plan and execute actions to accomplish corresponding tasks~\citep{anderson2018vision}. Recently, LLM-based large vision-language models (LVLMs) have demonstrated powerful capabilities in following instructions and performing planning based on the visual contexts~\citep{li2023blip,zhu2023minigpt,openai2023gpt4}, paving a promising path for the development of embodied AI systems.

However, recent studies have revealed significant deficiencies of LVLMs in understanding visual contents~\citep{li2023evaluating}. In terms of embodied scenarios, the ability to understand spatial relationships between objects is particularly vital for agents to effectively interact with the environment~\citep{anderson2018vision,padmakumar2022teach,li2022mvptr}, serving a wide range of embodied models~\citep{du2024delan, zhang2021curriculum, driess2023palm}. Evaluating and enhancing such capabilities of LVLMs is essential for constructing LVLM-driven embodied agents. Yet, existing benchmarks are not suitable for accurately assessing such capabilities. 


\begin{figure}[t]
    \vspace{0cm}
    \setlength{\abovecaptionskip}{0cm}
        \begin{center}
        \includegraphics[width=.48\textwidth]{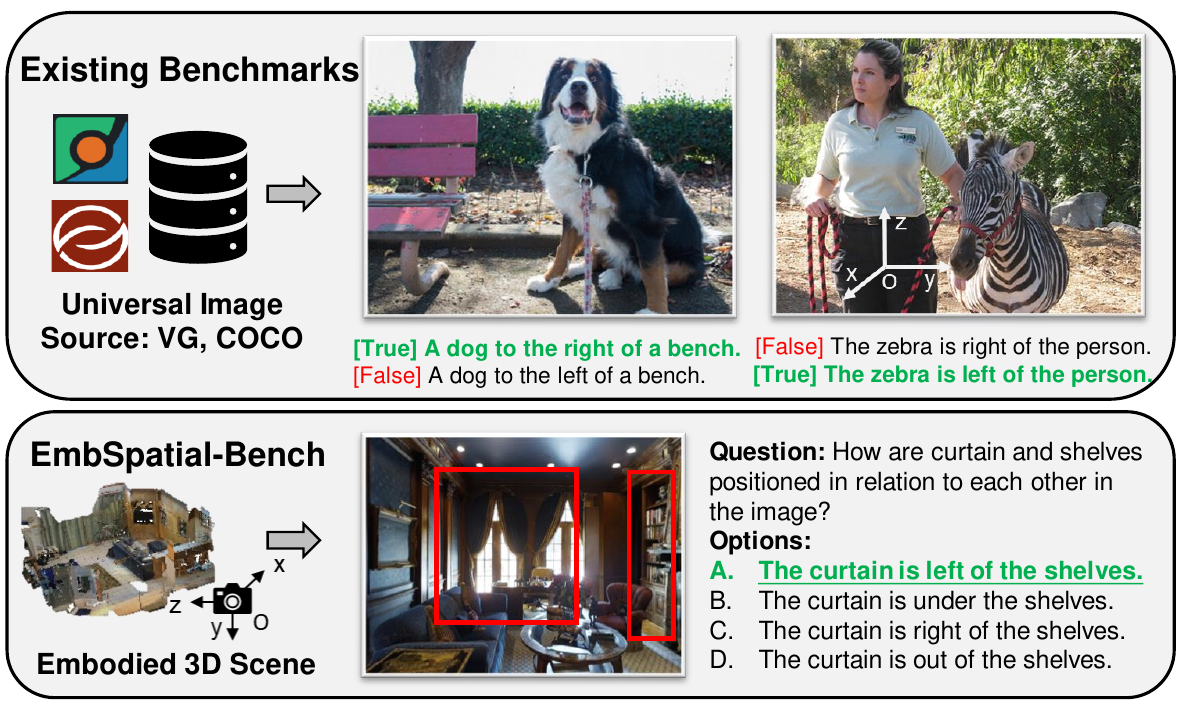}
        \end{center}
    \caption{Comparison between EmbSpatial-Bench and existing benchmarks for spatial understanding. Existing benchmarks may determine spatial relationships based on a coordinate system centered on the subject in the image (upper right), whereas EmbSpatial-Bench consistently determines them from an egocentric perspective.} 
    \label{fig:example1}
    \vspace{-0.5cm}
\end{figure}

In this paper, we argue that two important features should be considered for excellent evaluation of spatial understanding abilities in embodied tasks. 
First, the spatial relationships should be described from the egocentric perspective, for the reason that agents take themselves as the center of coordinates to follow instructions and infer decisions in embodied tasks. However, previous benchmarks for spatial understanding~\citep{liu2023visual} tend to depict spatial relationships from the perspective of subject within images, as illustrated in Figure~\ref{fig:example1}. Second, the visual scenes for evaluation should be consistent with that in embodied tasks. 
Nevertheless, existing benchmarks~\citep{liu2023visual,kamath2023s} are mainly constructed from universal image-text datasets like MSCOCO~\citep{lin2014microsoft} and VG~\citep{krishna2017visual} which are weakly related to embodied scenarios. 

\begin{figure*}[t]
    \vspace{0.0cm}
    \setlength{\abovecaptionskip}{0cm}
        \begin{center}
        \includegraphics[width=.9\textwidth]{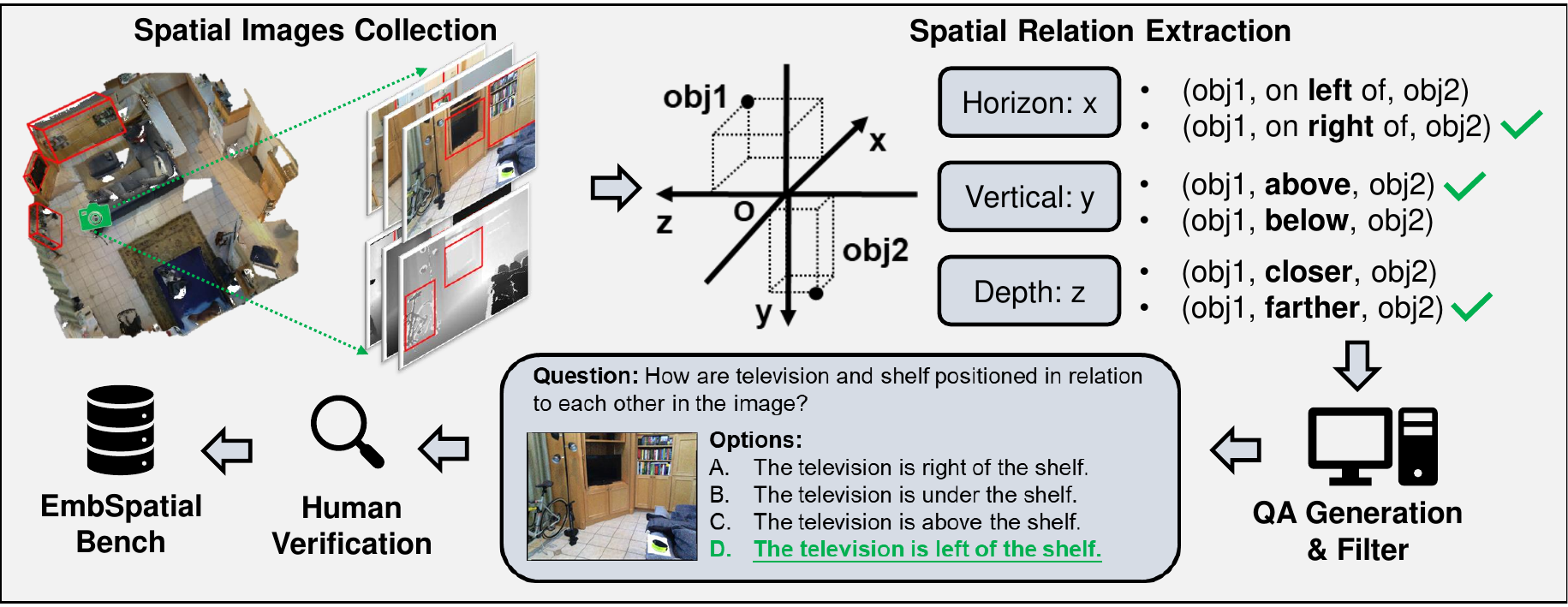}
        \end{center}
    \caption{Overview of the construction pipeline for EmbSpatial-Bench based on existing annotated 3D environments.} \label{fig:pipeline}
    \vspace{-0.5cm}
\end{figure*}

To meet aforementioned requirements, we establish EmbSpatial-Bench, a benchmark for evaluating spatial understanding abilities of LVLMs in embodied environments. As shown in Figure~\ref{fig:example1}, we focus on six spatial relationships described from the egocentric perspective, including \emph{above}, \emph{below}, \emph{left}, \emph{right}, \emph{close} and \emph{far}, which completely covers three dimensions of the coordinates. The benchmark is organized into the format of multiple-choice questions. The images used for evaluation are directly collected from embodied 3D scenes, namely MP3D~\citep{chang2017matterport3d}, AI2-THOR~\citep{kolve2017ai2} and ScanNet~\citep{dai2017scannet}. 

Based on EmbSpatial-Bench, various LVLMs have been assessed. Experimental results indicate the poor embodied spatial understanding of current LVLMs, including GPT-4V~\citep{openai2023gpt4} and Qwen-VL-Max~\citep{bai2023qwen}. 
To address the issue, we further construct an instruction-tuning dataset, EmbSpatial-SFT, to empower LVLMs with embodied spatial understanding ability.
LVLMs fine-tuned on EmbSpatial-SFT consistently demonstrate improved spatial perception abilities across different scenarios. \footnote{\url{https://github.com/mengfeidu/EmbSpatial-Bench}}

\section{EmbSpatial-Bench}
\label{embspatial-bench}
Unlike existing benchmarks built on 2D images~\citep{liu2023visual}, EmbSpatial-Bench is constructed from 3D scenes. Figure~\ref{fig:pipeline} illustrates the construction pipeline. We first generate target images from 3D scenes and extract spatial relations among objects. Then, we generate QA pairs and conduct filtering. Section~\ref{dataset_construction} provides detailed explanations of each part, while Section~\ref{dataset_stat} offers statistics of the benchmark.


\begin{figure*}[t]
    \vspace{0.0cm}
    \setlength{\abovecaptionskip}{-0.1cm}
    \begin{center}
    \includegraphics[width=.98\textwidth]{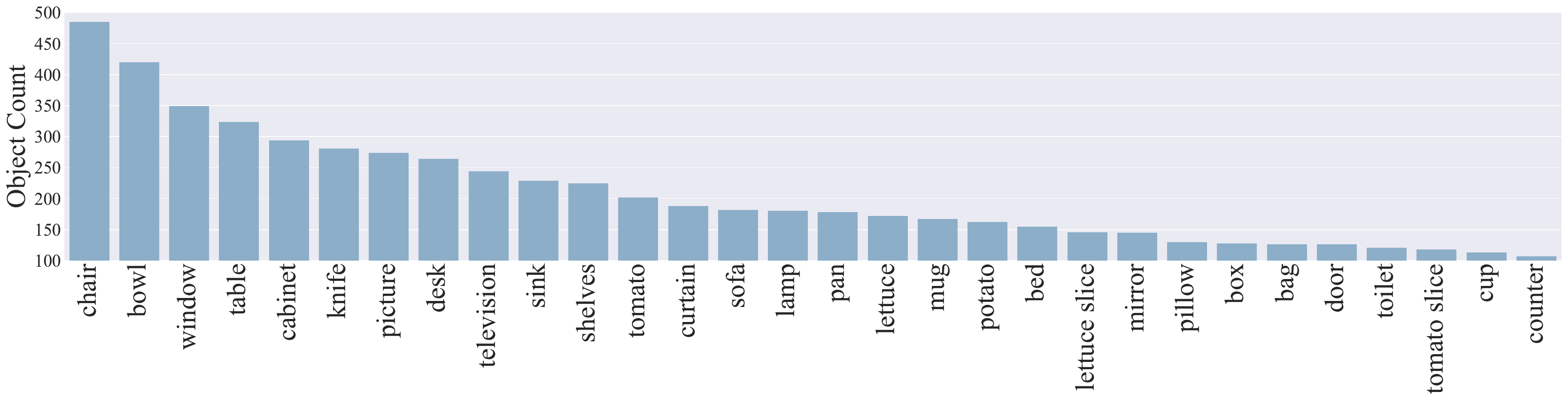}
    \end{center}
    \caption{Distribution of top 30 object categories.}
    \label{fig:top30 objects} 
\end{figure*}

\subsection{Dataset Construction}
\label{dataset_construction}

\paragraph{Spatial Image Sources.} 
Current embodied 3D simulators offer comprehensive annotations for tasks such as visual navigation~\citep{chang2017matterport3d} and room rearrangement~\citep{RoomR}, making them ideal for constructing a challenging benchmark to evaluate embodied spatial understanding. Therefore, we choose MP3D~\citep{chang2017matterport3d}, ScanNet~\citep{dai2017scannet} and AI2-THOR~\citep{kolve2017ai2}. 
Specifically, we utilize the test scenes from MP3D and validation scenes from ScanNet and A. Within each 3D scene, we randomly select viewpoints and capture the corresponding RGB-D images accordingly.
In AI2-THOR, we select 7 types of household tasks from ALFRED~\citep{shridhar2020alfred}, spanning 93 different scenes. During task execution, we identify key RGB-D images based on the dataset's PDDL~\citep{aeronautiques1998pddl} annotations.(See Appendix~\ref{sec:appendix_dataset}).

\paragraph{Spatial Relation Extraction.} 
Instead of relying on object detectors~\citep{gokhale2023benchmarking}, we extract spatial relations directly from well-annotated 3D datasets. 
For each object in each image, we can utilize the camera parameters along with the corresponding 3D coordinates to obtain its 2D coordinates in the image (in the form of bounding boxes).
With the 2D annotations, we extract the spatial relation triples with non-overlapping bounding boxes. We consider six spatial relationships from the viewer's perspective:  \emph{above}, \emph{below}, \emph{left}, \emph{right}, \emph{close} and \emph{far}. For the first four types, we determine the spatial relation based on position of the entire bounding boxes. 
For instance, if the entire bounding box of object A is located to the left of object B, we consider the relationship between A and B as \emph{A is left of B}. 
For the other two types, we use the average depth within the bounding box to determine which object is farther or closer.

\paragraph{QA Generation.} 
The format of our benchmark is multiple-choice questions, a widely adopted approach in various LVLM benchmarks~\citep{liu2023mmbench,li2023reform}. 
For the relations \emph{above}, \emph{below}, \emph{left} and \emph{right}, we design 5 templates to generate questions asking spatial relations between objects, with unrelated relations provided as false options.
For the relations \emph{far} and \emph{close}, we aggregate the relation triples for each image and generate questions for identifying the farthest or closest one among the given objects in the image.

\paragraph{Filtering and Human Verification.} 
To ensure the reliability of our benchmark, we initially filter out QA pairs with overly large or small bounding boxes, while maintaining a balanced distribution of spatial relations. Subsequently, we check each sample and remove the inappropriate questions that referring unclear objects or wrong spatial relationships. See appendix~\ref{sec:appendix_verify} for more filtering and human verification details.

\begin{table}[t]
\setlength{\abovecaptionskip}{0.15cm}
\setlength{\belowcaptionskip}{0cm}
    \centering
    \resizebox{.48\textwidth}{!}{
    \begin{tabular}{ccccc}
\toprule
\textbf{Data Source} & \textbf{\#QA Pairs}& \textbf{\#Image} & \textbf{\#Object}& \textbf{\#Scene} \\ \midrule 
Matterport3D      & 1,201     & 928     & 133 & 26\\
AI2-THOR      & 1,239     & 683     & 95 & 93\\ 
ScanNet     & 1,200    & 570     & 35 & 175\\ 
\midrule 
Overall      & 3,640     & 2,181     & 294 & 277\\ 
\bottomrule
\end{tabular}}
    \caption{Dataset Statistics of EmbSpatial-Bench}
    \label{tab:data_statistics}
    \vspace{-0.5cm}
\end{table}

\subsection{Dataset Statistics}
\label{dataset_stat}
\begin{figure}[t]
    \vspace{0cm}
    \setlength{\abovecaptionskip}{0cm}
    \begin{center}
    \includegraphics[width=.36\textwidth]{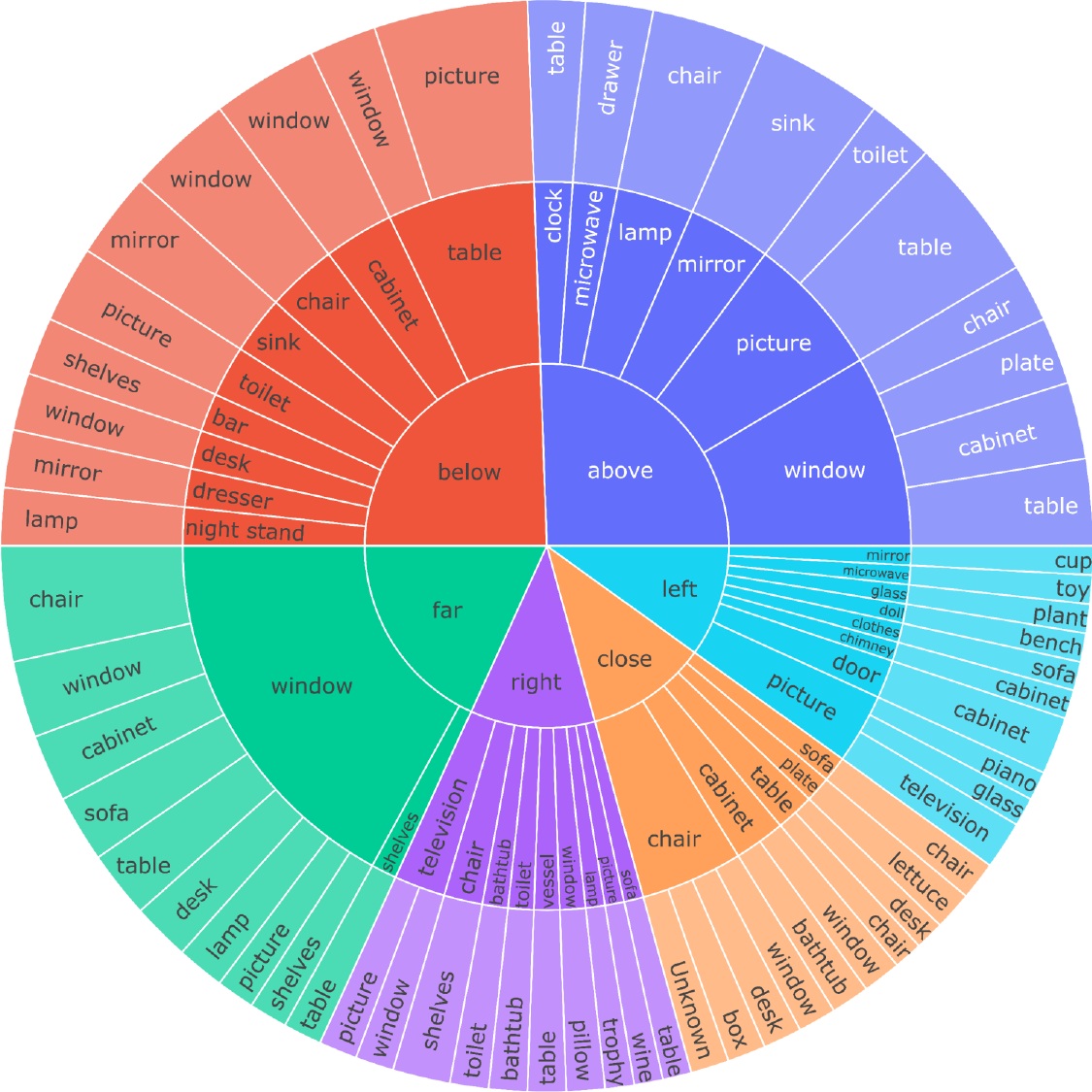}
    \end{center}
    \caption{The top 10 most common triples from each spatial relation in EmbSpatial-Bench.}
    \label{fig:pie_all} 
    \vspace{-0.5cm}
\end{figure}

As shown in Table~\ref{tab:data_statistics}, the constructed benchmark comprises a total of 3,640 QA pairs, 
covering 294 object categories and 6 spatial relationships. 
The distribution of top 30 object categories can be observed in Figure~\ref{fig:top30 objects}. The set of objects is the collection among samples from three embodied datasets. Indoor objects such as "chair", "bowl" and "window" are the most frequent across different scenes.
The distribution of most common spatial relation triples are depicted in Figure~\ref{fig:pie_all}, highlighting the diversity of the combination of object spatial relations present in our benchmark. We also maintain a balanced distribution of spatial relations (details in Appendix~\ref{sec:appendix_dataset}).
The diversity and balance of the data enhance the reliability of our benchmark.


\section{EmbSpatial-SFT}
\label{embspatial-SFT}

To further improve LVLMs' capacity in embodied spatial understanding, we construct an instruction-tuning dataset, EmbSpatial-SFT, which provides QA data for two tasks: spatial relationship identification and object localization. The former task setting is consistent with EmbSpatial-Bench, while the latter serves as an auxiliary task to enhance the model's ability to ground target objects. The auxiliary task can be considered as the foundational skill for relationship identification. EmbSpatial-SFT is solely built on the training split of MP3D. In this way, we can still conduct zero-shot evaluations of the instruction-tuned models using data from the other two scenes in EmbSpatial-Bench.

\paragraph{Spatial Relation Identification.}
Following the automatic pipeline in Section~\ref{embspatial-bench}, We construct 25K training samples for spatial relation identification.

\paragraph{Object Localization.} 
 Based on the coordinates of objects in 2D images, we construct object localization data in the form of the object grounding task~\citep{kazemzadeh2014referitgame}. The model is supposed to answer the location of inquired objects. The location is represented in the textual format of bounding boxes, following~\citet{chen2023minigpt}.


\section{Experiments}

\subsection{Experimental Setup}

Based on EmbSpatial-Bench, we conduct zero-shot evaluation of current LVLMs, using accuracy as the metric. Two evaluation strategies are employed. The first one is the generation-based strategy, which directly uses predicted options from the textual outputs of models. Considering the insufficient instruction-following ability of some LVLMs, we also employed a likelihood strategy, using the option with the highest probability generated by the model~\citep{li2023reform}. Please refer to Appendix~\ref{sec:appendix_experiment} for more evaluation details.

\subsection{Zero-shot Performance}

Table~\ref{tab:lvlms results} presents the zero-shot performance of 10 open-source LVLMs and 2 closed-source models. The results indicate that current LVLMs, including powerful closed-source models like GPT-4V and Qwen-VL-Max, have not demonstrated satisfactory spatial understanding abilities in embodied scenes. The best performance among all LVLMs merely reaches an accuracy of 49.11\% (Generation) or 43.85\% (Likelihood) which is significantly lower than human performance (90.33\%). 
We present failure cases of GPT-4V in Appendix~\ref{sec:appendix_gpt4v}, revealing its poor abilities of both object localization and spatial relation identification. The versions of these models can be found in Appendix~\ref{model_introduction}.

\begin{table}[t]
    \setlength{\abovecaptionskip}{0.15cm}
    \setlength{\belowcaptionskip}{0cm}
    \vspace{0cm}
        \centering
        \resizebox{.45\textwidth}{!}{
        \begin{tabular}{lcc}
    \toprule
    \textbf{Model} & \textbf{Generation}& \textbf{Likelihood}\\ \midrule 
    BLIP2~\citeyearpar{li2023blip}      & 37.99    & 35.71     \\
    InstructBLIP~\citeyearpar{dai2023instructblip}      & 38.85      & 33.41     \\
    Cheetor~\citeyearpar{li2023finetuning}      & 24.56     & 32.80    \\
    Lynx~\citeyearpar{zeng2023matters}     & 29.09    & 41.62     \\
    mPlugOwl~\citeyearpar{ye2023mplug}     & 24.12     & 27.42    \\
    ImagebindLLM~\citeyearpar{han2023imagebind}     & 26.46     & 33.46     \\
    Shikra~\citeyearpar{chen2023shikra}     & 28.38     & 34.75     \\
    MiniGPT4~\citeyearpar{zhu2023minigpt}      & 23.54     & 31.70     \\
    MiniGPT-v2~\citeyearpar{chen2023minigpt}      & 23.93     & \textbf{43.85}     \\ 
    LLaVA-1.6~\citeyearpar{liu2023improved}     & \textbf{35.19}     & 38.84     \\
    \midrule 
    GPT-4V~\citeyearpar{openai2023gpt4}      & 36.07    & -     \\ 
    Qwen-VL-Max~\citep{bai2023qwen}      & \textbf{49.11}    & -     \\ 
    \midrule 
    Human      & \textbf{90.33}    & -     \\ 
    \bottomrule
    \end{tabular}}
        \caption{Zero-shot performance (Acc\%) of LVLMs in EmbSpatial-Bench. \textbf{Bold} indicates the best results.}
        \label{tab:lvlms results}
        \vspace{-0.5cm}
\end{table}

\subsection{Instruction Tuning on EmbSpatial-SFT}

Furthermore, we fine-tune MiniGPT-v2 on EmbSpatial-SFT, to explore whether the data could further enhance the model's spatial understanding capabilities. The trainable parameters include the visual connection module and LoRA~\citep{hu2021lora} modules in the LLM backbone.

\paragraph{Main Results.}
According to Table~\ref{tab:finetune results}, under the likelihood evaluation strategy, learning from EmbSpatial-SFT consistently improves the performance across both in-domain and out-domain environments, with an increase of 34.25\% in the overall accuracy.
Though not as significant as that under likelihood strategy, the evaluated results under generation strategy still demonstrate an adequate performance improvement (+9.04\% overall) after instruction-tuning. 
The improvement in AI2-THOR is less than in ScanNet, which we attribute to AI2-THOR primarily consisting of simulated scenes, unlike the real-world scenarios in MP3D and ScanNet.

\paragraph{Ablations.}
We further validate the effectiveness of finetuning LLM backbone with LoRA and the auxiliary object localization data. As shown in Table~\ref{tab:finetune results}, tuning the LLM backbone with LoRA significantly contributes to the performance across all scenarios compared to the variant with a frozen LLM backbone. 
This phenomenon implies the necessity for the LLM backbone to learn corresponding reasoning abilities for spatial understanding, rather than solely adjusting the input visual representations.
The auxiliary data also contribute to the performance across different embodied environments, leading to an overall improvement of 0.47\% and 0.76\% under generation strategy and likelihood strategy, respectively.

    
    

\begin{table}[t]
     \vspace{0cm}
    \setlength{\abovecaptionskip}{0.15cm}
    \setlength{\belowcaptionskip}{0cm}
    \centering
    \resizebox{.48\textwidth}{!}{
    \begin{tabular}{lcccc}
    \toprule
    \multicolumn{1}{l|}{\multirow{2}{*}{\textbf{Model}}} & \textbf{In-Domain} & \multicolumn{2}{c|}{\textbf{Out-Domain}}                  & \multirow{2}{*}{\textbf{All}} \\ \cmidrule{2-4}
    \multicolumn{1}{l|}{}                                & \textbf{MP3D}      & \textbf{AI2-THOR} & \multicolumn{1}{c|}{\textbf{ScanNet}} &                               \\ \midrule
    \multicolumn{5}{c}{\textbf{Generation}}                                                                                                                               \\ \midrule
    \multicolumn{1}{l|}{MiniGPT-v2~\citeyearpar{chen2023minigpt}}                      & 23.31              & 20.58             & \multicolumn{1}{c|}{28.00}            & 23.93                         \\ \midrule
    \multicolumn{1}{l|}{Finetuned MiniGPT-v2}            & 31.64              & \textbf{34.06}    & \multicolumn{1}{c|}{\textbf{33.17}}   & \textbf{32.97}                \\
    \multicolumn{1}{l|}{\enspace w/o LoRA}                        & 26.81              & 25.26             & \multicolumn{1}{c|}{23.25}            & 25.11                         \\
    \multicolumn{1}{l|}{\enspace w/o OL}                          & \textbf{34.22}     & 31.40    & \multicolumn{1}{c|}{31.92}            & 32.50                         \\ \midrule
    \multicolumn{5}{c}{\textbf{Likelihood}}                                                                                                                               \\ \midrule
    \multicolumn{1}{l|}{MiniGPT-v2~\citeyearpar{chen2023minigpt}}                      & 46.71              & 41.97             & \multicolumn{1}{c|}{42.92}            & 43.85                         \\ \midrule
    \multicolumn{1}{l|}{Finetuned MiniGPT-v2}            & \textbf{80.52}     & \textbf{73.69}    & \multicolumn{1}{c|}{\textbf{80.25}}   & \textbf{78.10}                \\
    \multicolumn{1}{l|}{\enspace w/o LoRA}                        & 48.38              & 38.90             & \multicolumn{1}{c|}{44.17}            & 43.76                         \\
    \multicolumn{1}{l|}{\enspace w/o OL}                          & 80.35              & 72.15    & \multicolumn{1}{c|}{79.67}            & 77.34                         \\ \bottomrule
    \end{tabular}}
    \caption{Performance (Acc\%) of MiniGPT-v2 tuned on EmbSpatial-SFT. OL stands for object localization while w/o LoRA indicates that only the connection module is fine-tuned. \textbf{Bold} indicates the best results.}
    \label{tab:finetune results}
    \vspace{-0.5cm}
\end{table}

\section{Related Works}
\paragraph{Large Vision-Language Models}
The prevalent LVLMs~\citep{dai2023instructblip,zeng2023matters} learn visual representations from abundant image-text interleaved datasets with a lightweight connection module.
Further works~\citep{tsai2023multimodal,zheng2023towards} fine-tunes LVLMs-based architecture and obtain acceptable performance on embodied tasks, which preliminarily reveal the potential of LVLMs as embodied intelligence. However, these works neither evaluate nor empower LVLMs with spatial understanding ability, which is essential for various embodied tasks.

\paragraph{Benchmarks for Spatial Understanding.}

While there are numerous universal benchmarks available for LVLMs~\citep{xu2023lvlm,fu2023mme,li2023reform}, dedicated benchmarks for evaluating spatial understanding remain scarce.
VSR~\citep{liu2023visual} typically examines spatial relationships from the perspective of the subject within the image. 
What'sUp~\citep{kamath2023s} addresses data bias 
and generates uncluttered images to eliminate interference from unrelated objects.
$ \rm{SR_{2D}} $~\citep{gokhale2023benchmarking} focuses on evaluating text-to-image generative model.
However, all of them are built on COCO~\citep{veit2016coco} or VG~\citep{krishna2017visual} which are not consistent with the embodied scenarios. This lack of specialized benchmarks leaves the spatial understanding capabilities of LVLMs in embodied tasks unexplored.

\section{Conclusion}
In this work, we propose EmbSpatial-Bench, a benchmark to evaluate embodied spatial understanding of LVLMs.
The evaluation results reveal the weak spatial understanding ability of current popular LVLMs.
We further propose EmbSpatial-SFT, an instruction tuning dataset to enhance the capacity of LVLMs.
Extensive experiments valid the effectiveness of each data component in our EmbSpatial-SFT, with the goal of empowering the spatial understanding ability of LVLMs. 

\section*{Limitations}
Spatial understanding in embodied environments is a crucial aspect of LVLMs' capabilities for embodied tasks. In this study, we advance towards this goal by constructing benchmark and instruction-tuning datasets from well-annotated 3D embodied datasets. These datasets are derived from three widely used indoor embodied datasets, which may restrict their suitability for outdoor environments. Additionally, our study only investigates the English language, thus limiting the generalizability of the benchmark and findings to other languages.

\section*{Ethical Considerations}
The benchmark and instruction-tuning data are built from publicly available embodied datasets, which include either photorealistic scenes or generated rendered scenes without any copyright issues. Besides,
our data source does not contain any personal data, uniquely identifiable individuals, or offensive content.

\section*{Acknowledgements}
This work is supported by National Natural Science Foundation of China (No. 62176058) and National Key R\&D Program of China (2023YFF1204800). The project's computational resources are supported by  CFFF platform of Fudan University.


\bibliography{custom}

\newpage
\appendix
\begin{appendices}
\section{Dataset Details}
\label{sec:appendix_dataset}
\subsection{AI2-THOR Image Selection}
\label{sec:ai2thor_scene}
Due to the significant similarity between many images in the observation sequences for each task in AI2-THOR, filtering is necessary. Based on the detailed PDDL annotations from ALFRED~\citep{shridhar2020alfred}, we select key images that show significant content changes after each sub-goal is reached as our benchmark image resources.

\subsection{Dataset Statistics}
\begin{figure}[t]
    \vspace{0cm}
    \setlength{\abovecaptionskip}{-0.1cm}
    \begin{center}
    \includegraphics[width=.3\textwidth]{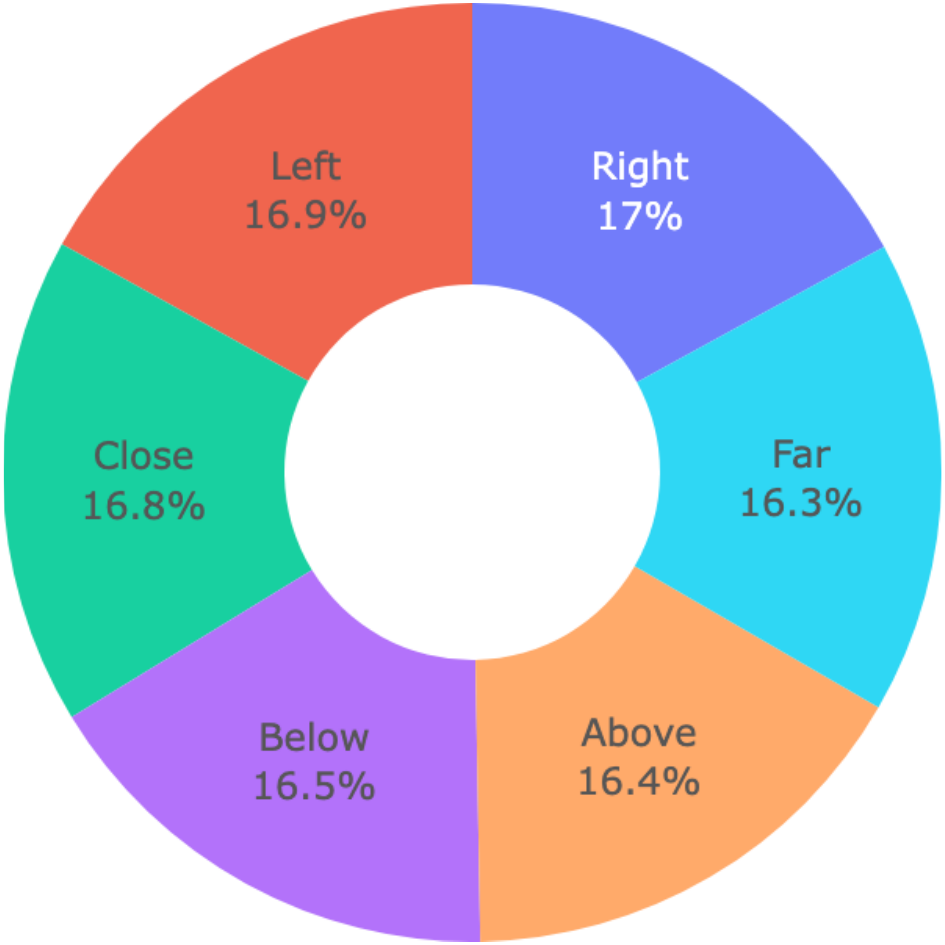}
    \end{center}
    \caption{Distribution of spatial relationships in EmbSpatial-Bench.}
    \label{fig:appendix_data_relation} 
\end{figure}

The wordcloud of object categories can be observed in Figure~\ref{fig:appendix_data_wordcloud}. The distribution of questions for each spatial relation is illustrated in Figure~\ref{fig:appendix_data_relation}. The diversity and balance of the data enhance to the reliability of our benchmark.

\subsection{Data Cases}
\label{sec:appendix_data_cases}
\begin{figure}[t]
    \vspace{0cm}
    \setlength{\abovecaptionskip}{0cm}
    \begin{center}
    \includegraphics[width=.48\textwidth]{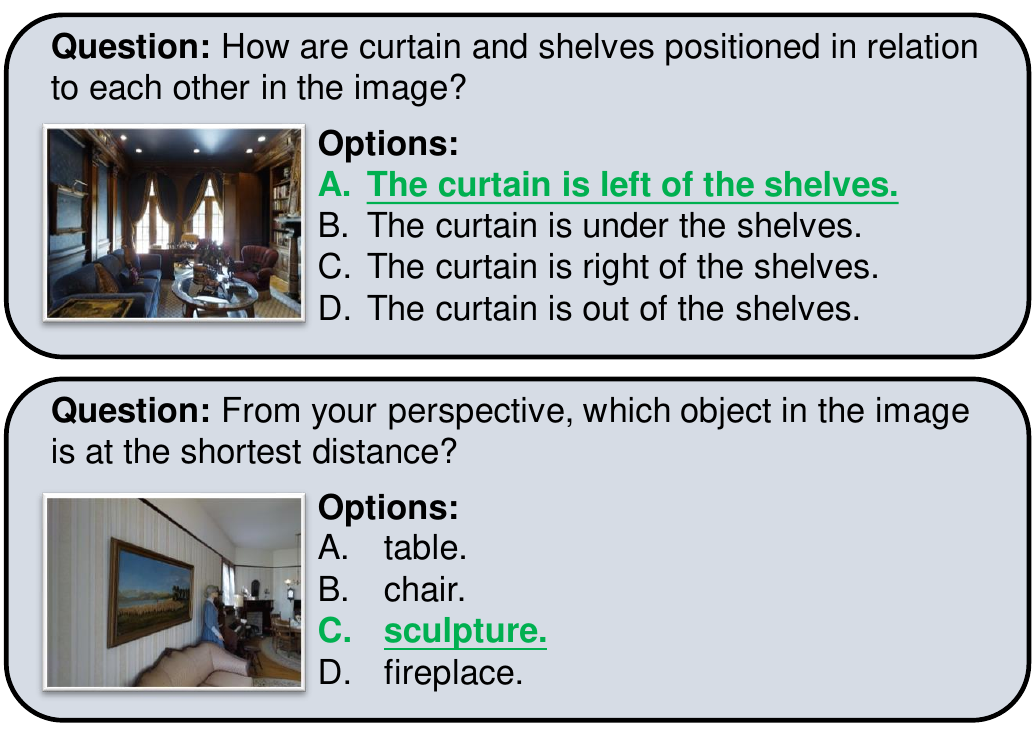}
    \end{center}
    \caption{Data samples from Matterport3D.} 
    \label{fig:mp3d_case}
    \vspace{0cm}
\end{figure}

\begin{figure}[t]
    \vspace{0cm}
    \setlength{\abovecaptionskip}{0cm}
    \begin{center}
    \includegraphics[width=.48\textwidth]{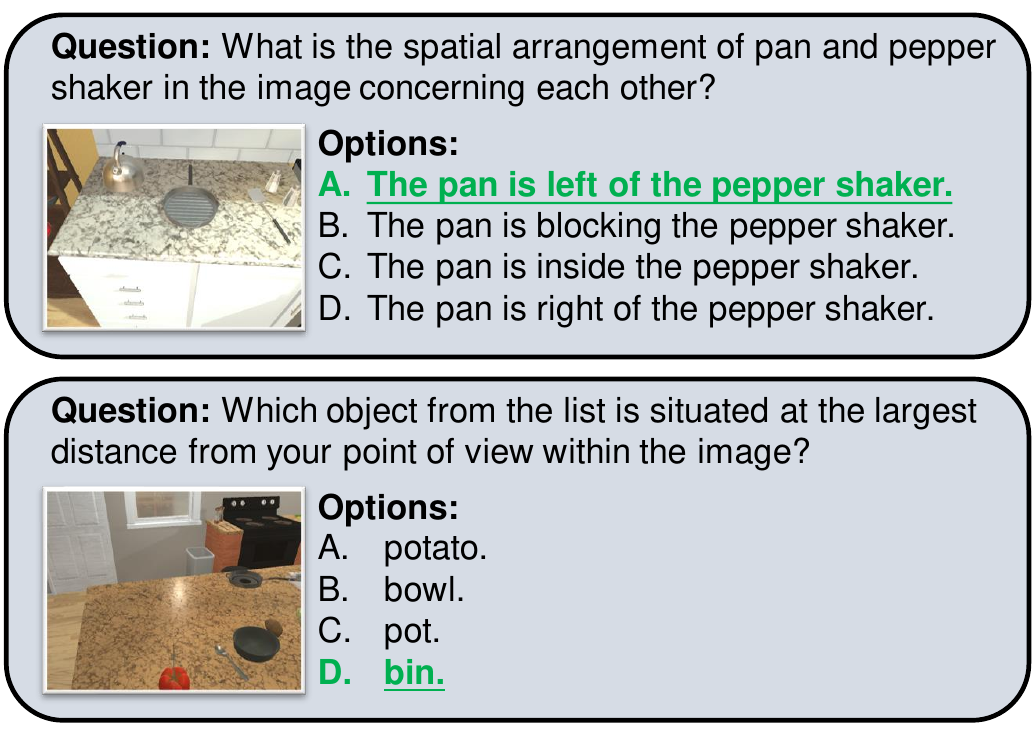}
    \end{center}
    \caption{Data samples from AI2-THOR.} 
    \label{fig:ai2thor_case}
    \vspace{0cm}
\end{figure}

\begin{figure}[t]
    \vspace{0cm}
    \setlength{\abovecaptionskip}{0cm}
    \begin{center}
    \includegraphics[width=.48\textwidth]{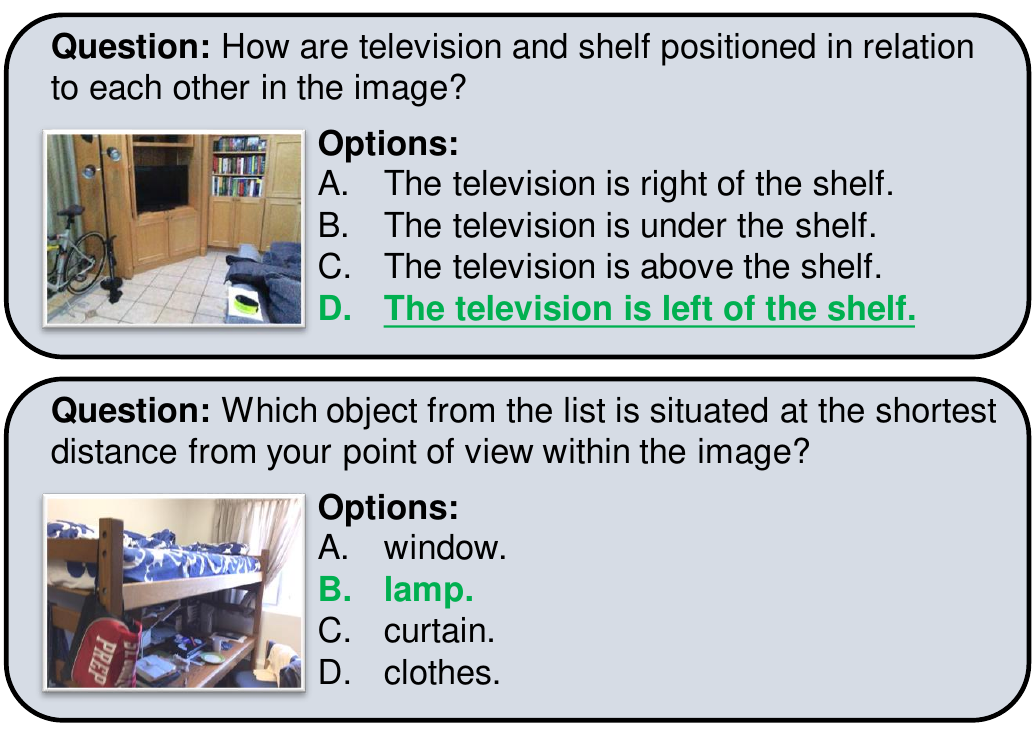}
    \end{center}
    \caption{Data samples from ScanNet.} 
    \label{fig:scannet_case}
    \vspace{0cm}
\end{figure}

Three samples of EmbSpatial-Bench constructed from MP3D~\citep{chang2017matterport3d}, AI2-THOR~\citep{kolve2017ai2} and ScanNet~\citep{dai2017scannet} are shown in Fig.~\ref{fig:mp3d_case}, Fig.~\ref{fig:ai2thor_case} and Fig.~\ref{fig:scannet_case}.

\subsection{Filtering and Verification}
\label{sec:appendix_verify}
Initially, we will implement two primary filtering processes to enhance the robustness and quality of our benchmark. First, we filter out objects with excessively large or small bounding boxes. To exclude improperly displayed objects, we filter out spatial relationship triplets where the length or width of the bounding box is less than 50 or greater than half the length of the corresponding dimension of the image.

After automated construction and filtering processes, the human verification is implemented to further ensure the correctness of our benchmark. Specifically, the correctness of each sample is examined by human from several aspects: 1) the objects involved in the question can be identified in the image uniquely and clearly; 2) the target object conforms to the described spatial relationship; 3) the negative options are indeed incorrect objects or relationships. Any sample that does not meet either of these conditions is discarded.

\begin{figure}[t]
    \vspace{0cm}
    \setlength{\abovecaptionskip}{-0.1cm}
    \begin{center}
    \includegraphics[width=.48\textwidth]{figs/wordcloud1.pdf}
    \end{center}
    \caption{Wordcloud of object categories.}
    \label{fig:appendix_data_wordcloud} 
\end{figure}

\section{Experiments}
\label{sec:appendix_experiment}
\subsection{Experimental Details}
\paragraph{Implementation details.} We use MiniGPT-v2~\citep{chen2023minigpt} as a baseline LVLM for investigation. The architecture of MiniGPT-v2 comprises three components, including a vision encoder , a linear connection layer and a large language model. We initialize the model parameters with the official checkpoint after its instruction-tuning. We finetune the connection layer and the large language model of MiniGPT-v2 with LoRA~\citep{hu2021lora}. In our implementation, we set the LoRA rank, $R_r$ = 64 and scaling factor, $R_\alpha$ = 16. 

\paragraph{Training and hyper-parameters. } We adopt AdamW optimizer with a cosine learning rate scheduler during the finetune process. The model is finetuned for 25,000 steps on 4xV100 GPUs with a initial learning rate of 1e-5, a minimum learning rate of 1e-6, a warmup learning rate of 1e-6 and a global batch size of 16. The finetuning stage lasts around 10 hours.

\subsection{Evaluation Strategy}
Following the evaluation approach~\citep{li2023reform}, we evaluate LVLMs with generation and likelihood strategy. 
The likelihood strategy relies on LVLMs' intrinsic nature as generative models and separates their instruction-following capacity from the capacity being evaluated. Given the image $v$, the question $q$, and $N$ options $C=\{c^i\}_{i=1}^{N}$, the prediction can be determined by the generation likelihood of LVLM: 
{\setlength{\abovedisplayskip}{0.1cm}
\setlength{\belowdisplayskip}{0cm}
\begin{align}
    \hat{c}=\arg\max_{c^i\in C}P_{\theta}(c^i|v,q)
\end{align}}
where $P_{\theta}(c^i|v,q)$ is parameterized by the causal-LLM-based LVLMs. The generation strategy extracts the option mark from generated textual output as predicted option.

\subsection{Models}
\label{model_introduction}
We select 10 open-source and 2 closed-source LVLMs for a comprehensive evaluation, including BLIP2~\citep{li2022blip}, InstructBLIP~\citep{dai2023instructblip}, Cheetor~\citep{li2023finetuning}, Lynx~\citep{zeng2023matters}, mPlugOwl~\citep{ye2023mplug}, ImagebindLLM~\citep{han2023imagebind}, Shikra~\citep{chen2023shikra}, MiniGPT4~\citep{zhu2023minigpt}, MiniGPT-v2~\citep{chen2023minigpt}, LLaVA-1.6~\citep{liu2023improved}, GPT-4V~\citep{openai2023gpt4}, Qwen-VL-Max~\citep{bai2023qwen}.
Among the open-source models, BLIP2 and InstructBLIP have the Flant5 LLM backbones. The LLM backbone of Cheetor, Lynx, MiniGPT4 and LLaVA1.6 is Vicuna~\citep{vicuna2023}. mPlugOwl chooses LLaMA~\citep{gao2023llama} as backbone and MiniGPTv2 choooses LLaMA2~\citep{touvron2023llama2} as backbone. All experimental open-source models have a parameter size of approximately 7B. We select version of ``gpt-4-1106-vision-preview'' for GPT-4V. 

\subsection{Main Results of Each Spatial Relation}
We have analyse the models' performance before and after instruct-tuning on different spatial relations, as shown in the table~\ref{tab:finetune results across relation}. 

After instruct-tuning on EmbSpatial-SFT, MiniGPT-v2 significantly improved or maintained comparable accuracy on various spatial relationship categories across different environments. In the likelihood evaluation, compared to the horizontal and vertical dimensions, performance in the depth dimension is significantly lower. We attribute this to the training data of LVLMs lacking depth estimation and the need to identify four objects in complex scenes, instead of just two objects in the other two dimensions. 
In the generation evaluation, both MiniGPT-v2 and the fine-tuned model perform poorly. Improving generation performance of open-source models remains an open question for further exploration.

\begin{table*}[!ht]
    \centering
    \resizebox{.98\textwidth}{!}{
    \begin{tabular}{ccccccccccccccccccc}
    \toprule
        \multicolumn{1}{c|}{\multirow{3}{*}{\textbf{Model}}} & \multicolumn{6}{c}{\textbf{In-Domain}} & \multicolumn{12}{c}{\textbf{Out-Domain}}\\ \cmidrule{2-19}
        \multicolumn{1}{c|}{} & \multicolumn{6}{c|}{\textbf{MP3D}} &\multicolumn{6}{c|}{\textbf{AI2THOR}} &\multicolumn{6}{c}{\textbf{ScanNet}}\\ 
        \multicolumn{1}{c|}{} & \textbf{above} & \textbf{below} & \textbf{left} & \textbf{right} & \textbf{close} & \multicolumn{1}{c|}{\textbf{far}} & \textbf{above} & \textbf{below} & \textbf{left} & \textbf{right} & \textbf{close} & \multicolumn{1}{c|}{\textbf{far}} & \textbf{above} & \textbf{below} & \textbf{left} & \textbf{right} & \textbf{close} & \textbf{far} \\ 
        \cmidrule{1-19}
        ~ & \multicolumn{18}{c}{\textbf{Generation}}\\ \cmidrule{1-19}
        \multicolumn{1}{c|}{MiniGPT-v2} & 31.22 & 26.90 & \textbf{24.76} & 20.48 & 21.29 & \multicolumn{1}{c|}{15.54} & 25.60 & 23.41 & 18.93 & 16.19 & 25.24 & \multicolumn{1}{c|}{13.93} & 31.00 & 33.00 & \textbf{30.00} & 22.50 & \textbf{29.50} & 22.00 \\ 
        \multicolumn{1}{c|}{Finetuned MiniGPT-v2} & \textbf{38.62} & \textbf{46.19} & 22.86 & \textbf{23.33} & \textbf{34.65} & \multicolumn{1}{c|}{\textbf{25.39}} & \textbf{36.23} & \textbf{49.76} & \textbf{28.16} & \textbf{26.67} & \textbf{34.76} & \multicolumn{1}{c|}{\textbf{28.86}} & \textbf{39.00} & \textbf{48.00} & 27.50 & \textbf{23.00} & 28.50 & \textbf{33.00} \\      
        \cmidrule{1-19}
        ~ & \multicolumn{18}{c}{\textbf{Likelihood}}\\ \cmidrule{1-19}
        \multicolumn{1}{c|}{MiniGPT-v2} & 91.01 & 76.65 & 30.95 & 30.48 & 25.74 & \multicolumn{1}{c|}{29.53} & 79.71 & 62.93 & 30.58 & 25.24 & 32.38 & \multicolumn{1}{c|}{20.9} & 78.50 & 73.50 & 28.00 & 32.50 & 27.00 & 18.00 \\ 
        \multicolumn{1}{c|}{Finetuned MiniGPT-v2} & \textbf{92.59} & \textbf{91.88} & \textbf{84.29} & \textbf{82.38} & \textbf{71.78} & \multicolumn{1}{c|}{\textbf{60.10}} & \textbf{93.72} & \textbf{88.78} & \textbf{83.50} & \textbf{80.95} & \textbf{50.00} & \multicolumn{1}{c|}{\textbf{44.77}} & \textbf{90.50} & \textbf{89.00} & \textbf{89.50} & \textbf{90.50} & \textbf{56.50} & \textbf{65.50} \\ \bottomrule
    \end{tabular}}
    \caption{Performance (Acc\%) of MiniGPT-v2 and fine-tuned MiniGPT-v2 across different spatial relations.}
\label{tab:finetune results across relation}
\end{table*}

    
    


\section{GPT-4V Cases}
\label{sec:appendix_gpt4v}
\begin{figure*}[t]
    \vspace{-16cm}
    \setlength{\abovecaptionskip}{0cm}
    \begin{center}
    \includegraphics[width=.98\textwidth]{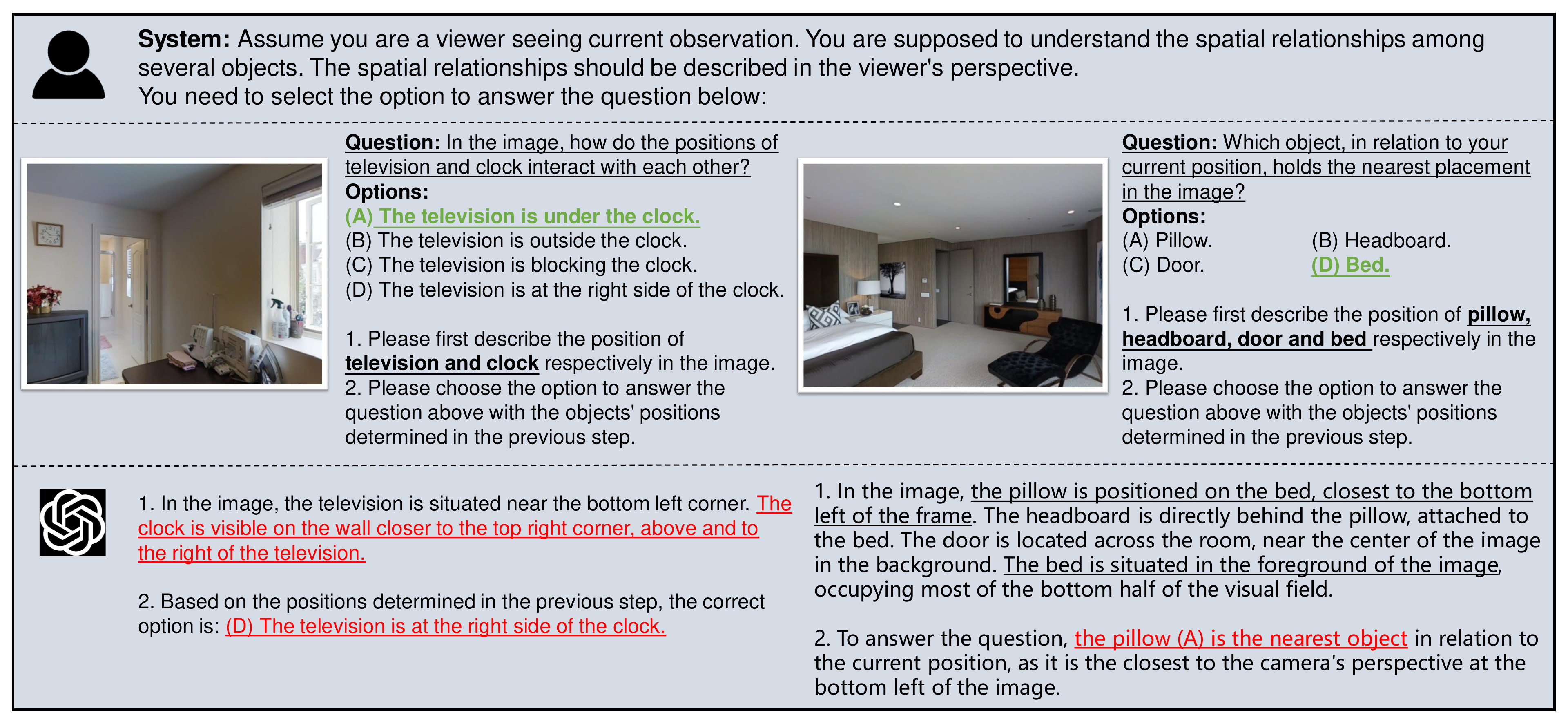}
    \end{center}
    \caption{Case study of GPT-4V on our benchmark. It not only makes mistakes in object positioning, but also misjudge their spatial relationship when successfully localizing the objects involved. \textcolor{red}{The text in red} means the wrong answers generated by GPT-4V.} 
    \label{fig:gpt4_case}
    \vspace{-0.5cm}
\end{figure*}

Utilizing the strong instruction following ability of GPT-4V, we delved deeper into the possible reasons for the poor performance of current LVLMs. Inspired by the two processes decoupling from spatial understanding, we prompt GPT-4V to inspect whether object localization or spatial relationships determination becomes a bottleneck. As shown in Figure~\ref{fig:gpt4_case}, the GPT-4V not only makes mistakes in object positioning, but also misjudge their spatial relationship when successfully localizing the objects involved. In the first case (left part), GPT-4V mistakenly positions the clock in top left corner to the top right corner, further leading to the incorrect selection of option with the word "right". In the second case (right part), GPT-4V successfully locates the positions of all object referred in the question, but incorrectly choose the pillow as the nearest object rather than the bed. The case study demonstrate the potential room for improvement in both two processes.

\end{appendices}
\end{document}